\documentclass[10pt,twocolumn,letterpaper]{article}
\usepackage[accsupp]{axessibility}  %

\usepackage[pagenumbers]{cvpr} 

\usepackage{booktabs}   
\usepackage{multirow}   
\usepackage{makecell}   
\usepackage{array}      
\usepackage{caption}    

\definecolor{cvprblue}{rgb}{0.21,0.49,0.74}
\usepackage[pagebackref,breaklinks,colorlinks,allcolors=cvprblue]{hyperref}
\usepackage[acronym, nohypertypes={acronym}]{glossaries}
\newacronym{AE}{AE}{autoencoder}
\newacronym{VAE}{VAE}{variational autoencoder}
\newacronym{DM}{DM}{diffusion model}
\newacronym{pgd}{PGD}{Projected Gradient Descent}
\newacronym{cs}{CS}{Cosine Similarity}

\usepackage{balance}
\usepackage{dblfloatfix}
\usepackage{textcase}
\usepackage{amsmath}
\usepackage{amsmath,amsfonts,bm}

\def\eqref#1{equation~\ref{#1}}

\def\1{\bm{1}}

\def\ve{{\bm{e}}}

\def\vx{{\bm{x}}}

\DeclareMathAlphabet{\mathsfit}{\encodingdefault}{\sfdefault}{m}{sl}
\SetMathAlphabet{\mathsfit}{bold}{\encodingdefault}{\sfdefault}{bx}{n}

\def\paperID{}
\def\confName{CVPR}
\def\confYear{2026}

\title{Breaking the Illusion: Consensus-Based Generative  Mitigation of Adversarial Illusions in Multi-Modal Embeddings}

\author{Fatemeh Akbarian\textsuperscript{ ‡ ,*}, Anahita Baninajjar\textsuperscript{ ‡ ,*}, Yingyi Zhang\textsuperscript{ ‡ ,}\thanks{  Equal contribution. \textsuperscript{†}Contributed in an advisory role.} , Ananth Balashankar\textsuperscript{†}, Amir Aminifar\textsuperscript{ ‡}\\
\small\textsuperscript{‡}Lund University, Sweden \textsuperscript{†}Google DeepMind, USA\\
{\tt\small\{fatemeh.akbarian, anahita.baninajjar, yingyi.zhang, amir.aminifar\}@eit.lth.se,}\\
{\tt\small ananthbshankar@google.com}\\
}

\begin{document}

\maketitle

\begin{abstract}
Multi-modal foundation models align images, text, and other modalities in a shared embedding space but remain vulnerable to adversarial illusions \cite{bagdasaryan2024adversarial}, where imperceptible perturbations disrupt cross-modal alignment and mislead downstream tasks. To counteract the effects of adversarial illusions, we propose a task-agnostic mitigation mechanism that purifies the attacker's perturbed input using generative models, e.g., Variational Autoencoders (VAEs), to restore natural alignment. To further enhance the defense mechanism, we adopt a generative sampling strategy combined with a consensus-based aggregation scheme over the outcomes of the generated samples. Our experiments on ImageBind, a state-of-the-art multi-modal encoder, show that our approach substantially reduces the illusion attack success rates to near-zero and improves cross-modal alignment in unperturbed and perturbed input settings, providing an effective and task-agnostic defense against adversarial illusions. The code is available at 
\href{https://github.com/fatemehakb/adversarial-illusions-mitigation}{https://github.com/fatemehakb/adversarial-illusions-mitigation}.
\end{abstract}  

\section{Introduction}

Multi-modal foundation models have rapidly advanced the frontier of visual and linguistic understanding. Foundation models such as CLIP~\cite{radford2021learning}, ALIGN~\cite{jia2021scaling}, and ImageBind~\cite{girdhar2023imagebind} align a variety of heterogeneous modalities including images and text within a shared embedding space, thereby enabling zero-shot classification, cross-modal retrieval, and generative conditioning.

The shared embedding space that underpins cross-modal flexibility simultaneously introduces a new attack space, giving rise to \emph{adversarial illusions}~\cite{bagdasaryan2024adversarial}. As downstream tasks directly rely on the integrity of this shared representation, even small perturbations in one modality can induce semantic misalignment across others, misleading models that depend on the embedding for retrieval, captioning, or generative conditioning. Defending against such cross-modal attacks presents unique challenges. Conventional adversarial defenses, originally developed for unimodal classifiers, emphasize local decision-boundary robustness~\cite{goodfellow2014explaining,madry2017towards,tramer2017ensemble,athalye2018obfuscated}. However, these techniques fail to maintain alignment in the shared embedding space. Even if accuracy is preserved, the perturbed features may still project to semantically inconsistent regions in the shared embedding space~\cite{zhou2023advclip,xia2025adversarial,bagdasaryan2024adversarial}. This misalignment undermines downstream tasks, where maintaining cross-modal coherence is essential.

Existing defenses against adversarial perturbations face unique challenges in multi-modal settings. Traditional preprocessing methods such as feature squeezing and JPEG compression~\cite{xu2017feature,guo2017countering} can suppress pixel-level noise but do not inherently restore cross-modal alignment. While adversarial fine-tuning schemes~\cite{schlarmann2024robust} successfully improve the resilience of the vision encoder, they still demand computationally expensive fine-tuning of the foundational vision components. To circumvent the need for retraining altogether, recent research has explored post-hoc, task-agnostic defenses, such as diffusion-based cumulative purification~\cite{fu2025diffcap}. However, relying exclusively on iterative diffusion processes during inference introduces substantial computational overhead. Thus, there remains a critical need for efficient, lightweight mitigation mechanisms capable of restoring cross-modal embedding alignment against adversarial illusions without incurring prohibitive inference costs.

In this paper, we propose a post-hoc task-agnostic mitigation mechanism against adversarial illusions. Our proposed mitigation processes adversarially perturbed inputs through a generative model such as \glspl{AE}, \glspl{VAE}, or \glspl{DM} to purify the attacker's perturbed input. By purifying inputs on a learned data manifold, these models suppress adversarial perturbations and project them back onto the natural cross-modal alignment. This generative purification framework mitigates adversarial illusions, restoring semantic consistency across modalities and preserving coherence in cross-modal generation.

To further enhance the mitigation mechanism, we adopt a generative sampling strategy combined with a consensus-based aggregation scheme over the outcomes of the generated samples. Generative sampling is the process of drawing multiple input purifications from a generative model, e.g., a \gls{VAE}, conditioned on a given input. 
This generative sampling and aggregation scheme, further enhances robustness by leveraging the diversity of stochastic purifications to mitigate adversarial illusions and restore cross-modal alignment. The result is a post-hoc task-agnostic defense that is both model-independent and scalable across modalities. 

We perform extensive experiments to demonstrate the effectiveness of our mitigation mechanism. First, our proposed mitigation improves the classification accuracy of the downstream task even when there is no adversarial illusions, from $42\%/66\%$ to $46\%/73\%$ in Top-1/Top-5 accuracies, respectively. Second, our proposed mitigation effectively neutralizes the attacker’s attempt to alter the downstream task’s prediction toward the targeted label, reducing the attack success rate from $62\%/90\%$ in Top-1/Top-5 accuracies to only $0\%/2\%$, respectively. 
Finally, our proposed mitigation not only effectively counteracts the attacker’s effort to steer the downstream task’s prediction toward the targeted label, but also realigns the perturbed input with the correct/original label, outperforming the state of the art, with Top-1/Top-5 accuracies improved from $32\%/56\%$ (for the best mitigation in \cite{bagdasaryan2024adversarial}) to $43\%/68\%$, respectively. 
We summarize our main contributions as follows:
\begin{itemize}
    \item We propose an effective mitigation mechanism against adversarial illusions in multi-modal embeddings based on a generative sampling strategy (using \glspl{AE}, \glspl{VAE}, or \glspl{DM}) combined with a consensus-based aggregation scheme over the generated samples (using majority voting). 
    \item We extensively evaluate our proposed mitigation mechanism against the state of the art and demonstrate that our proposed mitigation effectively restore cross-modal alignment under adversarial illusion attacks. For perturbed images, the cosine similarity with the correct text labels improves from $0.11 \pm 0.05$ (no mitigation) to $0.82 \pm 0.07$, while the Top-1/Top-5 accuracies increases from $0\%/0\%$ to $43\%/68\%$, substantially reducing adversarial illusion success rates without requiring retraining.
\end{itemize}

\section{Mitigation Framework}
In this section, we begin by detailing the threat model and explaining how the attacker manipulates cross-modal representations to induce semantic misalignment in adversarial illusion attack. Then, we introduce the goal of mitigating adversarial illusion attacks, while retaining cross-modal alignment on unperturbed inputs. We then describe our \textit{consensus-based generative mitigation}, which achieves this goal by projecting perturbed inputs back toward the natural data manifold before inference.

\subsection{Threat Model}
Adversarial illusion~\cite{bagdasaryan2024adversarial} is a targeted multi-modal attack where imperceptible perturbations are applied to an input to corrupt its genuine cross-modal alignment. By manipulating the shared embedding space, the attacker can steer any input to align closely with an arbitrary target sample from another modality, thereby inducing consistent semantic misalignment across downstream tasks. 

\begin{figure*}[t]
\centering
\includegraphics[trim={0 3.9cm 0 3cm},clip, width=1\textwidth]{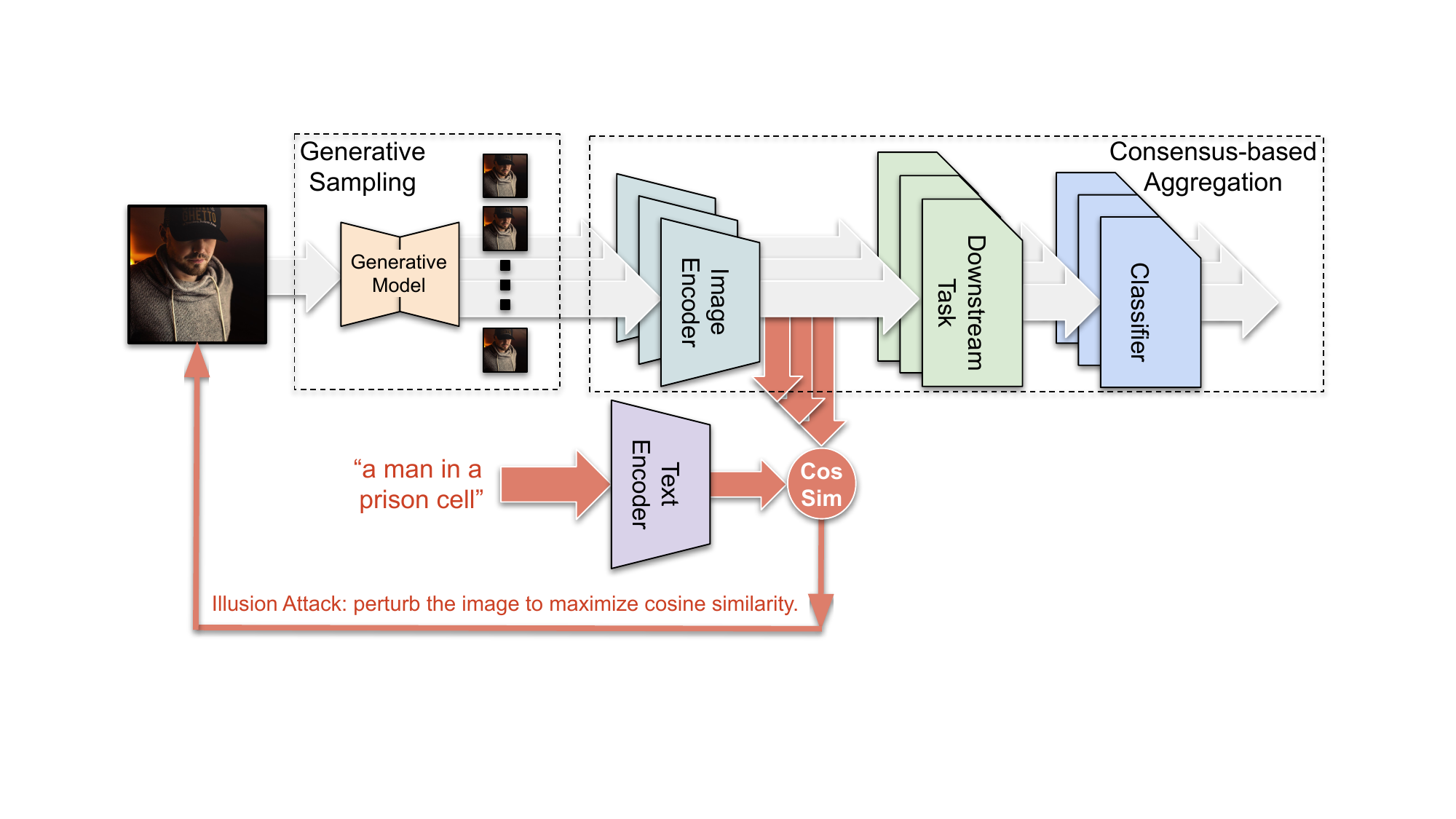}
\caption{Overview of our consensus-based generative sampling mitigation framework. Our mitigation scheme has two main components: a generative sampling and a consensus-based aggregation. The generative sampling mechanism adopts a generative model to purify several variants of the input image. The consensus-based aggregation mechanism aggregates the decisions for the generated samples, e.g., based on majority voting.  The red arrows describe the illusion attack, where the adversary attempts to maximize the cosine similarity between the embedding of the perturbed image and the embedding of the target text (i.e., \emph{``a man in a prison cell''} in this example).  }
\label{fig:method}
\end{figure*}

The overview of the adversarial illusion attack is shown in Figure \ref{fig:method}. The adversarial illusion attack is achieved through an iterative optimization process (red arrows in Figure \ref{fig:method}) using \gls{pgd}, which updates the perturbation to maximize the \gls{cs} between the embeddings of the perturbed input image and the designated target text (i.e., \emph{``a man in a prison cell''} in this example). This formulation makes the adversarial illusion task-agnostic, as it operates directly in the shared embedding space rather than relying on any specific downstream task. The adversarial illusion explicitly targets the embedding space of multi-modal models, exploiting the interchangeability of representations across modalities to induce cross-domain misalignment. Moreover, it exhibits strong transferability where illusions generated using one encoder remain effective across models with different encoders.

\subsection{Consensus-Based Generative Mitigation}

\textbf{Mitigation Goal}: Given clean or unperturbed inputs, and perturbed inputs generated to induce adversarial illusions in multi-modal embeddings; our goal is to two-part: \textit{(1) Minimize adversarial illusion attack success rates;  (2) Retain cross-modal alignment on unperturbed inputs; in a task agnostic manner.}

This formulation ensures that the mitigation approach improves robustness to adversarial illusion attacks without compromising the utility of the multi-modal models in benign settings. Further, achieving this goal in a task-agnostic defense ensures that a mitigation approach can scale seamlessly to various downstream applications. 

To achieve this goal, we leverage how adversarial illusions operate by driving the input representation toward an incorrect target in the shared embedding space. Our defense focuses on restoring the original alignment without altering the encoder or retraining downstream models. To this end, we propose a consensus-based generative sampling framework that purifies adversarially perturbed inputs prior to inference, aiming to recover representations consistent with the natural cross-modal structure. As illustrated in Figure~\ref{fig:method}, the framework combines generative purification with stochastic sampling to guide perturbed inputs back toward the natural data manifold. This purification step removes the adversarial shift toward the attacker’s target and maintains the prediction that the clean input would have produced, ensuring that the intended semantic label is preserved.

To ensure that our approach is task agnostic, we build our defense on top of pretrained generative models, rather than training a new purifier, because these models already approximate the natural image manifold and are widely available in multi-modal pipelines. They allow us to purify inputs with minimal pipeline changes as they naturally suppress off-manifold perturbations that adversarial illusions rely on. Using an off-the-shelf generator as a projection step lets us wrap the encoder with a lightweight defense, even when we only have black-box access to the downstream model. 

We now detail how the generative model transforms an adversarial input into a purified one. Given an adversarial input $\tilde{\vx} = \vx + \boldsymbol{\delta}$, where $\boldsymbol{\delta}$ denotes a small imperceptible perturbation crafted to disrupt alignment, our objective is to recover a purified input $\hat{\vx}$ whose representation preserves the semantics of the original input $\vx$. A generative model $\mathcal{G}_{\theta}$, e.g., \gls{AE}, \gls{VAE}, or \gls{DM}, acts as a post-hoc purifier that maps adversarially perturbed samples to purified inputs, where \( \hat{\vx}= \mathcal{G}_{\theta}(\tilde\vx). \) This purification maps the adversarial input into a latent or diffusion space, where adversarial perturbations are isolated and suppressed, and decodes it back to the natural input domain. Thus, $\mathcal{G}_{\theta}$ provides a smooth approximation of the underlying data manifold, suppressing high-frequency, off-manifold perturbations that induce misalignment in the shared embedding space.

Beyond deterministic purification, we leverage the stochastic nature of generative models to enable further robustness through sampling and aggregation. Given an adversarial input $\tilde\vx$, we draw $N$ independent purified samples such that
\[
{\hat{\vx}}_i = \mathcal{G}_{\theta}(\tilde\vx, \epsilon_i), \quad \epsilon_i \sim \mathcal{N}(0, \sigma^2 I),
\]
where $\epsilon_i$ denotes stochastic noise applied in the latent or diffusion space. Each purified sample ${\hat{\vx}}_i$ is evaluated by the downstream model to obtain a corresponding prediction $y_i = h(f({\hat{\vx}}_i))$, where $f$ is the multi-modal encoder and $h$ the task-specific head, e.g., classifier or retrieval function. The final prediction is obtained via majority aggregation \(
\hat{y} = \operatorname{mode}\big(\{y_i\}_{i=1}^{N}\big),
\) where $\operatorname{mode}$ selects the label that appears most frequently among the purified samples.

This consensus-based aggregation scheme leverages the diversity of generative purified samples to reduce sensitivity to localized perturbations in input space. Our defense further benefits from the inductive bias of generative models, which constrain purified samples to the natural data manifold. By integrating purification, stochastic sampling, and majority aggregation, the method effectively smooths the decision boundary in the shared embedding space, reducing the effective gradient available to the attacker. Consequently, small perturbations in input space no longer yield consistent alignment shifts across purified samples, significantly enhancing robustness to adversarial illusions. Our proposed approach achieves the defense goal as it is post-hoc and task-agnostic, requiring no modification to the target encoder or downstream network.

\section{Implementation Setup}

We evaluate our methods using the adversarial image–text pairs released with the Adversarial Illusions benchmark~\cite{bagdasaryan2024adversarial}. The dataset provides both \textit{original} and \textit{perturbed} images generated to induce targeted cross-modal misalignment in the \textbf{ImageBind} multi-modal embedding model. Each perturbed image is designed to align with an adversary-chosen target text while remaining visually similar to the original image.

\noindent\textbf{Datasets:}
The benchmark is derived from ImageNet categories and includes paired original and perturbed images with their associated original and target labels. This structure enables controlled comparison of semantic alignment recovery across different mitigation methods.

\noindent\textbf{Metrics:}
We report \acrfull{cs} between image embeddings $f(\vx)$ and label embeddings $\ve_y$ ($\uparrow$ is better for original label, $\downarrow$ for target label), which captures how closely the two representations align, defined as follows,
\begin{equation*}
\text{CS}(\vx,y)=\frac{f(\vx)^\top \ve_y}{\lVert f(\vx)\rVert\,\lVert \ve_y\rVert}.
\end{equation*}
We also evaluate Top–1 and Top–5 accuracy. Given model scores $s(\vx)$, the Top–$k$ set is defined as: 
$\text{Topk}(\vx,k)=\operatorname{argsort}(s(\vx))_{1:k}.$
The correctness indicator is
$\mathbf{1}_{\text{Top}k}(\vx,y)=\mathbb{I}[\,y\in\text{Topk}(\vx,k)\,].$
Top–1 reflects whether the highest scoring label is correct, and Top–5 checks if the correct label appears in the top five predictions, as follows,
\begin{align*}
\text{Top-1}=\tfrac{1}{N}\sum_i \mathbf{1}_{\text{Top}1}(\vx_i,y_i), \ \
\text{Top-5}=\tfrac{1}{N}\sum_i \mathbf{1}_{\text{Top}5}(\vx_i,y_i).
\end{align*}
“Original label” accuracy measures alignment on the clean or unperturbed inputs ($\uparrow$ is better), and “Target label” accuracy quantifies the misalignment with the adversary-specified target i.e. the \textit{adversarial illusion attack success rate} ($\downarrow$ is better).

\noindent\textbf{Generative-based purification:}
We adopt three pretrained generative models with different levels of capacity to purify inputs and suppress adversarial artifacts. The \textbf{\acrfull{AE}} performs deterministic projection that removes high-frequency perturbations through latent-space compression. The \textbf{\acrfull{VAE}} introduces stochastic sampling in the latent space, enabling diverse purifications around the natural image manifold. The \textbf{\acrfull{DM}} further enhances purification fidelity through iterative denoising, effectively restoring semantic content from perturbed inputs. All models are publicly available and used without fine-tuning. These generative priors map adversarial examples back onto the natural image manifold before re-encoding with ImageBind.

\section{Results}

\subsection{Generative Mitigation Performance} 

Table~\ref{table1} reports the performance of our proposed purification-based mitigation method employing various generative models against adversarial illusion effects on the ImageBind multi-modal embedding model. For the baseline \textbf{without mitigation}, the original image achieves $85\%$ Top-1 and $99\%$ Top-5 accuracy with a \gls{cs} of $1$, confirming that clean samples are correctly classified by the downstream model. On the target label, these original clean images achieve only $0\%$ Top-1 and $1\%$ Top-5 accuracy, with a \gls{cs} of $0.09 \pm 0.05$, indicating that the model does not naturally align them with the attacker-chosen target. The original reconstructed sample, obtained by passing the image through ImageBind and then through BindDiffusion as a downstream reconstruction module, shows reduced performance with $42\%$ Top-1 and $66\%$ Top-5 accuracy, indicating that the reconstruction process introduces loss of class-specific details even in the absence of an attack. The perturbed image, which differs from the original only by an imperceptible perturbation, still reaches $77\%$ Top-1 and $95\%$ Top-5 accuracy on the original label, but its \gls{cs} with the clean embedding drops to $0.11 \pm 0.05$, showing that the perturbation significantly distorts the latent representation while leaving the image visually unchanged. On the target label, the same perturbed images obtain $0\%$ for both Top-1 and Top-5 accuracies, with a \gls{cs} of $0.72 \pm 0.08$, indicating that the attack successfully shifts the embedding toward the adversary’s target even though the classifier does not yet predict it. Finally, the perturbed reconstructed sample, generated from the perturbed image via ImageBind and BindDiffusion, loses alignment with the true class ($0\%$ Top-1, $0\%$ Top-5) and instead aligns with the attacker's target (Top-1 = $62\%$, Top-5 = $90\%$), demonstrating that the illusion attack effectively transfers through the generative process and manipulates cross-modal semantics.

We observe that \textbf{\gls{AE}, \gls{DM}, and \gls{VAE}} mitigations are unable to achieve sub-goal of retaining original label accuracy compared to the baseline without mitigation,  even though they achieve the sub-goal of minimizing the target label accuracy on perturbed images and improving the \gls{cs} with the correct text label embeddings through purification.

\begin{table}[t]
\centering
\caption{Evaluation of generative mitigation against adversarial illusion on ImageBind multi-modal embedding model. Top-1/Top-5 accuracy and \gls{cs} are reported for original (Org.) and perturbed (Prt.) images and their corresponding reconstructions (Rec.), generated by the downstream task, on original and target labels.}
\resizebox{\columnwidth}{!}{
\begin{tabular}{cc ccc ccc}
\toprule
\multicolumn{2}{c}{} & \multicolumn{3}{c}{\textbf{Original label}} & \multicolumn{3}{c}{\textbf{Target label}} \\
\cmidrule(lr){3-5} \cmidrule(lr){6-8}
\multicolumn{1}{c}{\textbf{Method}} & \multicolumn{1}{c}{\textbf{Org./Prt.}} & Top-1 & Top-5 & CS & Top-1 & Top-5 & CS \\
\midrule

\multirow{4}{*}{\makecell[c]{Without \\Mitigation}} 
  & Org. Img. & $85\%$ & $99\%$ & $1$ & $0\%$  & $1\%$ & $0.09 \pm 0.05$ \\
  & Org. Rec. & $42\%$ & $66\%$ & - & $0\%$ & $1\%$ & - \\ \cline{2-8}
  & Prt. Img. & $77\%$ & $95\%$ & $0.11 \pm 0.05$ & $0\%$ & $0\%$ & $0.72 \pm 0.08$ \\
  & Prt. Rec. & $0\%$ & $0\%$ & - & $62\%$ & $90\%$  & - \\ 
\cmidrule(lr){1-8}

\multirow{4}{*}{\gls{AE}} 
  & Org. Img. & $65\%$ & $89\%$ & $0.67 \pm 0.09$ & $0\%$ & $0\%$ & $ 0.09 \pm 0.05$ \\
  & Org. Rec. & $18\%$ & $39\%$ & - & $0\%$ & $1\%$ & - \\ \cline{2-8}
  & Prt. Img. & $62\%$ & $82\%$ & $0.62 \pm 0.10$ & $0\%$ & $1\%$ & $0.10 \pm 0.05$ \\
  & Prt. Rec. & $18\%$ & $35\%$ & - & $0\%$ & $0\%$ & - \\ 
\cmidrule(lr){1-8}

\multirow{4}{*}{\gls{DM}} 
  & Org. Img. & $81\%$ & $98\%$ & $0.91 \pm 0.04$ & $0\%$ & $1\%$ & $0.09 \pm 0.05$ \\
  & Org. Rec. & $39\%$ & $64\%$ & -  & $0\%$ & $1\%$ & - \\ \cline{2-8}
  & Prt. Img. & $77\%$ & $95\%$ & $0.81 \pm 0.08$ & $0\%$ & $0\%$ & $0.11 \pm 0.05$ \\
  & Prt. Rec. & $25\%$ & $53\%$ & - & $0\%$ & $1\%$ & - \\ 
\cmidrule(lr){1-8}

\multirow{4}{*}{\gls{VAE}} 
  & Org. Img. & $84\%$ & $98\%$ & $0.93 \pm 0.02 $ & $0\%$ & $1\%$ & $0.08 \pm 0.05$ \\
  & Org. Rec. &  $43\%$ & $67\%$ & - & $0\%$ & $1\%$ & - \\ \cline{2-8}
  & Prt. Img. & $78\%$ & $96\%$ & $0.82 \pm 0.07$ & $0\%$ & $1\%$ & $0.12 \pm 0.05$ \\
  & Prt. Rec. & $22\%$ & $47\%$ & - & $0\%$ & $2\%$ & - \\ 
\cmidrule(lr){1-8}

\multirow{4}{*}{\makecell[c]{Ours \\ (\gls{DM} + \\Sampling)}}
  & Org. Img. & $80\%$ & $98\%$ & $0.92 \pm 0.04$ & $0\%$ & $1\%$ & $0.09 \pm 0.05$ \\
   & Org. Rec. & $45\%$ &$\boldsymbol{73\%}$ & - & $0\%$ & $1\%$ & - \\ \cline{2-8}
  & Prt. Img. & $77\%$ & $96\%$ & $0.81 \pm 0.08$ & $0\%$ & $1\%$ & $0.11 \pm 0.05$ \\
  & Prt. Rec. & $34\%$ & $61\%$ & - &  $0\%$ & $3\%$ & - \\  
\cmidrule(lr){1-8}

  \multirow{4}{*}{\makecell[c]{Ours \\ (\gls{VAE} + \\Sampling)}}
  & Org. Img. & $84\%$ & $98\%$ & $0.93 \pm 0.02$ & $0\%$ & $1\%$ & $0.08 \pm 0.05$ \\
  & Org. Rec. & $\boldsymbol{46\%}$ & $\boldsymbol{73\%}$ & - & $0\%$ & $2\%$ & - \\ \cline{2-8}
  & Prt. Img. & $78\%$ & $96\%$ & $0.82 \pm 0.07$ & 0\% & 1\% &  $0.12 \pm 0.05$ \\
  & Prt. Rec. & $\boldsymbol{43\%}$ & $\boldsymbol{68\%}$ & - & $0\%$ & $2\%$ & - \\

\bottomrule
\end{tabular}
} \label{table1}
\end{table}

\begin{figure}[t]
\centering
\includegraphics[width=0.85\columnwidth,trim={0 0 0 0},clip]{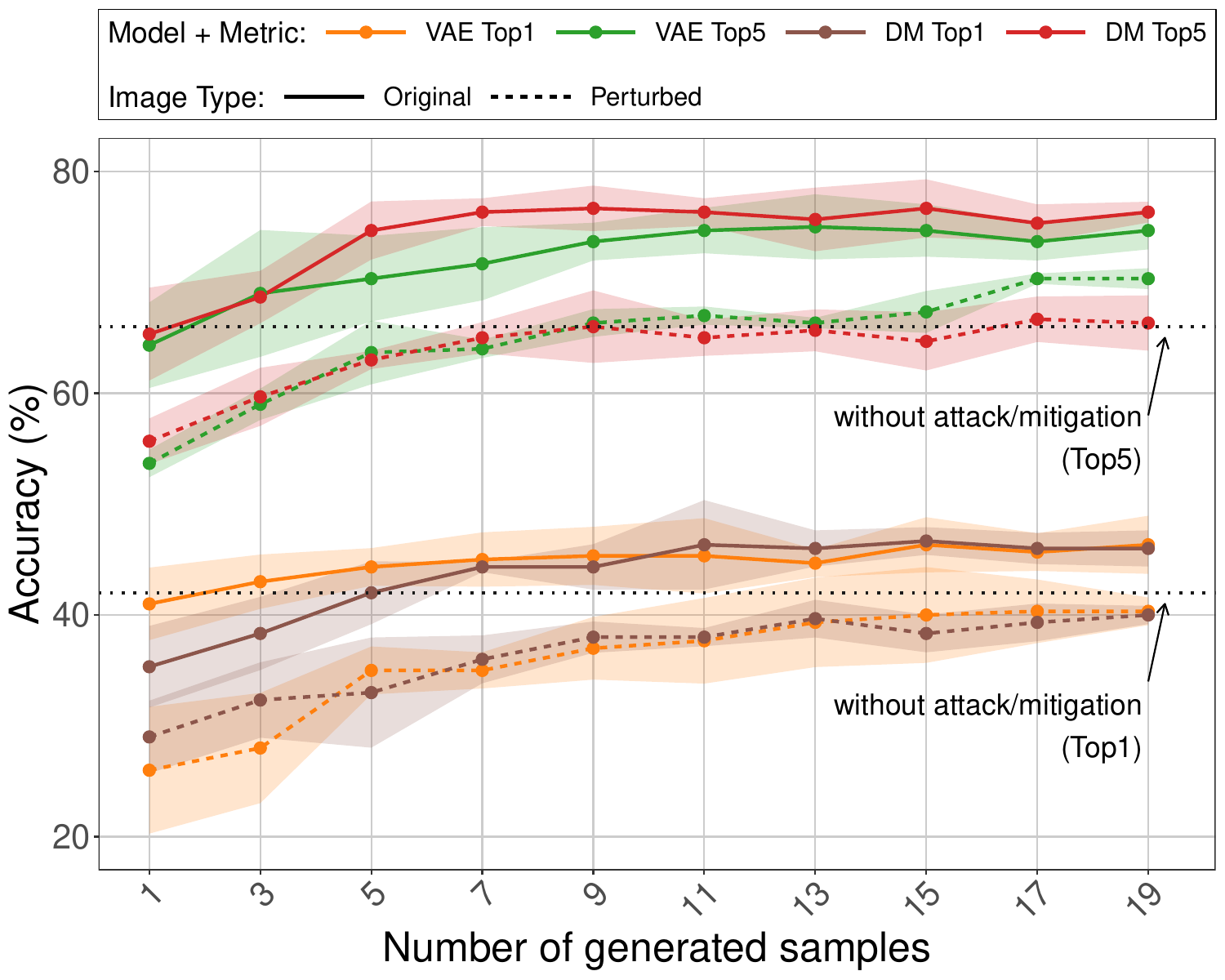}
\caption{Effect of sampling size for \gls{VAE} and \gls{DM} mitigation methods. Dotted lines show baseline without attack/mitigation for Top-1/Top-5 accuracy. Increasing the number of generated samples during the generative sampling stage improves both Top-1 and Top-5 accuracy for original and perturbed images, with performance stabilizing beyond $10$ generated samples.}
\label{sampling_effect}
\end{figure}

To evaluate our consensus-based generative sampling mitigation, we draw 10 samples per input from the VAE and DM. This choice is supported by Figure~\ref{sampling_effect}, which shows that both Top-1 and Top-5 accuracy plateau beyond roughly ten samples. Using $10$ samples, therefore, balances robustness and computational cost while capturing the full benefit of consensus-based generative sampling.
We find that \textbf{\gls{DM} + Sampling} achieves strong robustness on adversarial illusion attacks and stable cross-modal alignment on original and perturbed inputs. Finally, we find that our \textbf{\gls{VAE} + Sampling} configuration achieves the best overall purification performance. The original image reaches $84\%$ Top-1 and $98\%$ Top-5 accuracy with a \gls{cs} of $0.93 \pm 0.02$. The reconstructed image maintains 46\% Top-1 and 73\% Top-5 accuracy, exceeding the baseline for the same case, where no mitigation or attack is applied. This indicates that the proposed method does not introduce adverse effects under normal, non-adversarial conditions and can potentially improve performance. Under attack, the perturbed image achieves $78\%$ Top-1 and $96\%$ Top-5 accuracy with a \gls{cs} of $0.82 \pm 0.07$, while the reconstructed perturbed sample restores $43\%$ Top-1 and $68\%$ Top-5 accuracy. For comparison, in the normal setting without mitigation, the reconstructed image from the original image achieves $42\%$ Top-1 and $66\%$ Top-5 accuracy. This indicates that VAE + Sampling mitigation can even surpass the accuracy and performance of the normal setting under non-adversarial conditions.
For the target label, \gls{cs} remains very low at $0.08 \pm 0.05$ for the original image and $0.12 \pm 0.05$ for the perturbed image, and target-label accuracy stays near zero across all settings. These results show that adding stochastic sampling to the \gls{VAE} further improves robustness by suppressing adversarial bias and encouraging smoother, reliable purification.

\begin{table}[t]
\centering
\caption{Comparison of our best-performing configuration, i.e., \gls{VAE} + Sampling, to the mitigation strategies introduced in \cite{bagdasaryan2024adversarial} under the adversarial illusion attack on the ImageBind multi-modal embedding model. We report Top-1/Top-5 accuracy and \gls{cs} for original (Org.) and perturbed (Prt.) images and their reconstructions (Rec.), generated by the downstream task, on original and target labels.
}
\resizebox{\columnwidth}{!}{
\begin{tabular}{cc ccc ccc}
\toprule
\multicolumn{2}{c}{} & \multicolumn{3}{c}{\textbf{Original label}} & \multicolumn{3}{c}{\textbf{Target label}} \\
\cmidrule(lr){3-5} \cmidrule(lr){6-8}
\multicolumn{1}{c}{\textbf{Method}} & \multicolumn{1}{c}{\textbf{Org./Prt.}} & Top-1 & Top-5 & CS & Top-1 & Top-5 & CS \\
\midrule

\multirow{4}{*}{\makecell[c]{Without \\Mitigation}} 
  & Org. Img. & $85\%$ & $99\%$ & $1$ & $0\%$  & $1\%$ & $0.09 \pm 0.05$ \\
  & Org. Rec. & $42\%$ & $66\%$ & - & $0\%$ & $1\%$ & - \\ \cline{2-8}
  & Prt. Img.  & $77\%$ & $95\%$ & $0.11 \pm 0.05$ & $0\%$ & $0\%$ & $0.72 \pm 0.08$ \\
  & Prt. Rec. & $0\%$ & $0\%$ & - & $62\%$ & $90\%$  & - \\ 
\cmidrule(lr){1-8}

\multirow{4}{*}{JPEG} 
  & Org. Img. & $84\%$ & $98\%$ & $0.91 \pm 0.04$ & $0\%$ & $1\%$ & $0.09 \pm 0.05$ \\
  & Org. Rec. &  $37\%$ & $55\%$ & - & $0\%$ & $1\%$ & - \\ \cline{2-8}
  & Prt. Img.  & $79\%$ & $93\%$ & $0.78 \pm 0.07$ & $0\%$ & $1\%$ & $0.12 \pm 0.05$ \\ 
  & Prt. Rec. & $25\%$ & $50\%$ & - & $0\%$ & $2\%$ & - \\ 
\cmidrule(lr){1-8}

\multirow{4}{*}{\makecell[c]{Gaussian\\Blurred}}
  & Org. Img. & $70\%$ & $89\%$ & $0.77 \pm 0.05$ & $0\%$ & $1\%$ & $0.09 \pm 0.04 $\\
  & Org. Rec. & $29\%$ & $50\%$ & - & $0\%$ & $2\%$ & - \\ \cline{2-8}
  & Prt. Img.  & $65\%$ & $87\%$ & $0.74 \pm 0.07$ & $0\%$ & $0\%$ & $0.10 \pm 0.04$ \\
  & Prt. Rec. & $26\%$ & $53\%$ & - & $0\%$ & $0\%$ & - \\ 
\cmidrule(lr){1-8}

\multirow{4}{*}{\makecell[c]{Random\\Affine}}
  & Org. Img. & $83\%$ & $96\%$ & $0.84 \pm 0.05$ & $0\%$ & $0\%$ & $0.10 \pm 0.05$ \\
  & Org. Rec. & $36\%$ & $61\%$ & - & $0\%$ & $0\%$ & - \\ \cline{2-8}
  & Prt. Img.  & $81\%$ & $96\%$ & $0.78 \pm 0.07$ & $0\%$ & $0\%$ & $0.12 \pm 0.04$ \\
  & Prt. Rec. & $32\%$ & $56\%$ & - & $0\%$ & $1\%$ & - \\ 
\cmidrule(lr){1-8}

\multirow{4}{*}{\makecell[c]{Color\\Jitter}}
  & Org. Img. & $83\%$ & $99\%$ & $0.88 \pm 0.04$ & $0\%$ & $1\%$ & $0.09 \pm 0.05$ \\ 
  & Org. Rec. & $35\%$ & $60\%$ & - & $0\%$ & $1\%$ & - \\ \cline{2-8}
  & Prt. Img.  & $76\%$ & $95\%$ & $0.60 \pm 0.12$ & $0\%$ & $0\%$ & $0.22 \pm 0.06$ \\
  & Prt. Rec. & $12\%$ & $31\%$ & - & $1\%$ & $7\%$ & - \\ 
\cmidrule(lr){1-8}

\multirow{4}{*}{\makecell[c]{Horizontal\\Flip}}
  & Org. Img. & $84\%$ & $99\%$ & $0.90 \pm 0.03$ & $0\%$ & $1\%$ & $0.09 \pm 0.05$ \\
  & Org. Rec. & $45\%$ & $66\%$ & - & $0\%$ & $1\%$ & - \\ \cline{2-8}
  & Prt. Img.  & $77\%$ & $96\% $& $0.66 \pm 0.15$ & $0\%$ & $1\%$ & $0.20 \pm 0.08$ \\
  & Prt. Rec. & $17\%$ & $39\%$ & - & $0\%$ & $2\%$ & - \\ 
\cmidrule(lr){1-8}

\multirow{4}{*}{\makecell[c]{Random\\Perspective}}
  & Org. Img. & $72\%$ & $94\%$ & $ 0.80 \pm 0.06$ & $0\%$ & $0\%$ & $0.10 \pm 0.05$ \\ 
  & Org. Rec. & $26\%$ & $54\%$ & - & $0\%$ & $1\%$ & - \\ \cline{2-8}
  & Prt. Img.  & $74\%$ & $89\%$ & $0.73 \pm 0.09$ & $0\%$ & $0\%$ & $0.12 \pm 0.04$  \\ 
  & Prt. Rec. & $25\%$ & $52\%$ & - & $0\%$ & $2\%$ & - \\ 
\cmidrule(lr){1-8}

\multirow{4}{*}{\makecell[c]{Ours \\ (\gls{VAE} +\\ Sampling)}}
  & Org. Img. & $84\%$ & $98\%$ & $0.93 \pm 0.02$ & $0\%$ & $1\%$ & $0.08 \pm 0.05$  \\
  & Org. Rec. & $\boldsymbol{46\%}$ & $\boldsymbol{73\%}$ & - & $0\%$ & $2\%$ & - \\ \cline{2-8}
  & Prt. Img.  & $78\%$ & $96\%$ & $0.82 \pm 0.07$ & $0\%$ & $1\%$ & $0.12 \pm 0.05$ \\ 
  & Prt. Rec. & $\boldsymbol{43\%}$ & $\boldsymbol{68\%}$ & - & $0\%$ & $2\%$ & - \\ 
\bottomrule
\end{tabular}
} \label{table2}
\end{table}

\subsection{Comparison with Other Mitigation Methods} 

Table~\ref{table2} compares our proposed generative purification mitigation method with several standard mitigation techniques, including JPEG compression, Gaussian blur, random affine transformation, color jitter, horizontal flip, and random perspective. These methods operate directly in pixel space and are evaluated under the same adversarial illusion setup as our generative method.

State-of-the-art data augmentation methods provide only limited protection against adversarial illusions. JPEG compression and Gaussian blurring offer slight improvements, recovering $25\%$ and $26\%$ Top-1 accuracy on perturbed reconstructions, but they also reduce original reconstruction accuracy to $37\%$ and $29\%$, respectively. Other augmentations, such as random affine, color jitter, horizontal flip, and random perspective, show similar behavior. They partly disrupt the adversarial perturbation while introducing substantial semantic degradation. For example, random affine recovers $32\%$ Top-1 accuracy on reconstructed perturbed inputs, and color jitter and random perspective yield only $12\%$ and $25\%$. \gls{cs} to the original label drops sharply under these transformations, for instance $0.60 \pm 0.12$ for color jitter and $0.66 \pm 0.15$ for horizontal flip. In addition, color jitter produces elevated \gls{cs} to the target label up to $0.22 \pm 0.06$, indicating residual adversarial influence.
In contrast, our \textbf{\gls{VAE} + Sampling} defense achieves substantially higher recovery, restoring $43\%$ Top-1 and $68\%$ Top-5 accuracy on perturbed reconstructions while preserving strong clean-image performance and maintaining near-zero target-label alignment $0.12 \pm 0.05$. Unlike deterministic pixel-level transformations, the generative model explicitly projects inputs back onto the natural data manifold, suppressing adversarial noise without removing class-relevant structure. These results show that generative mitigation provides a more principled and semantically grounded defense strategy that yields significantly stronger robustness.

\begin{figure*}[t]
\centering

\begin{minipage}{0.32\textwidth}
    \centering
    \includegraphics[width=\linewidth]{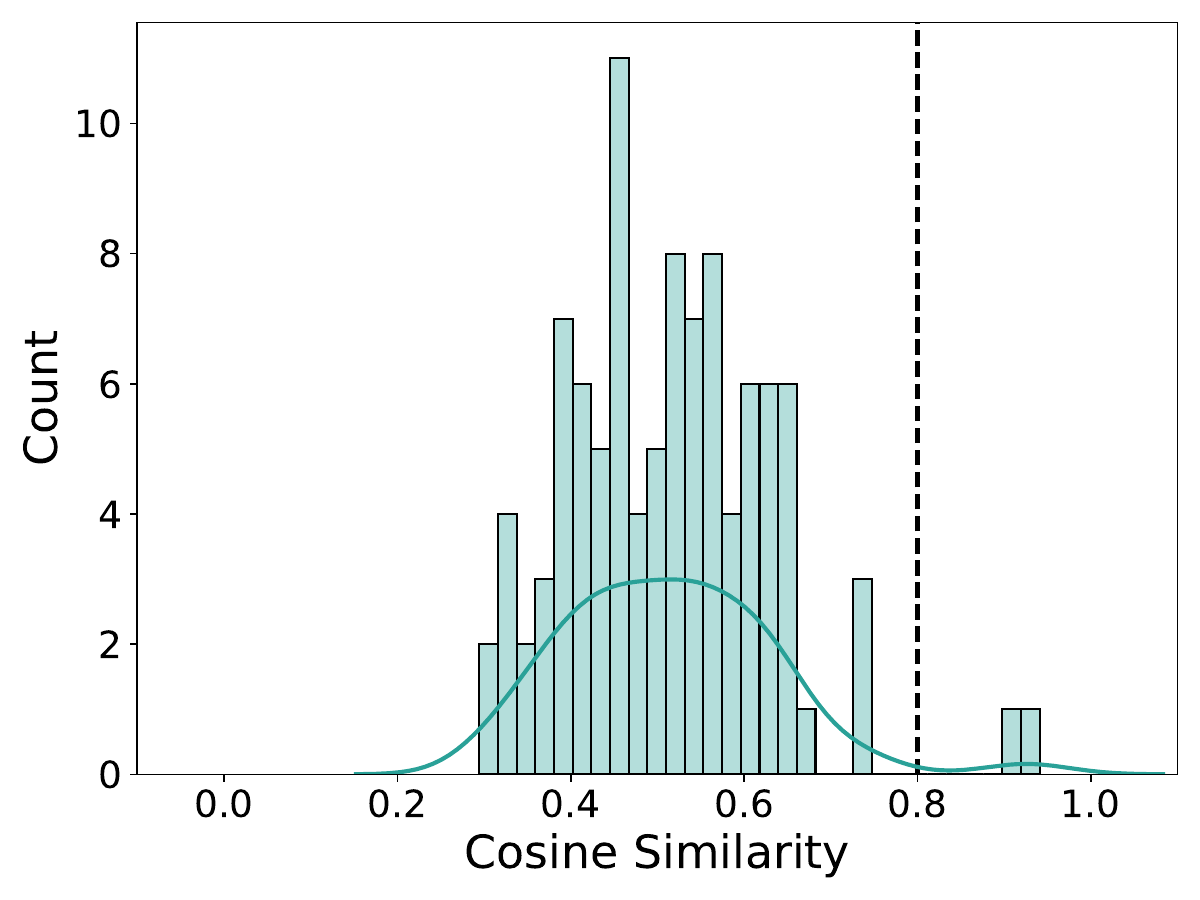}
    \caption{Distribution of cosine similarities between perturbed embeddings and target labels across attack iterations. These results guide the selection of a \gls{cs} threshold of $0.8$.}
    \label{cosine_exp}
\end{minipage}
\hfill
\begin{minipage}{0.32\textwidth}
    \centering
    \includegraphics[width=\linewidth]{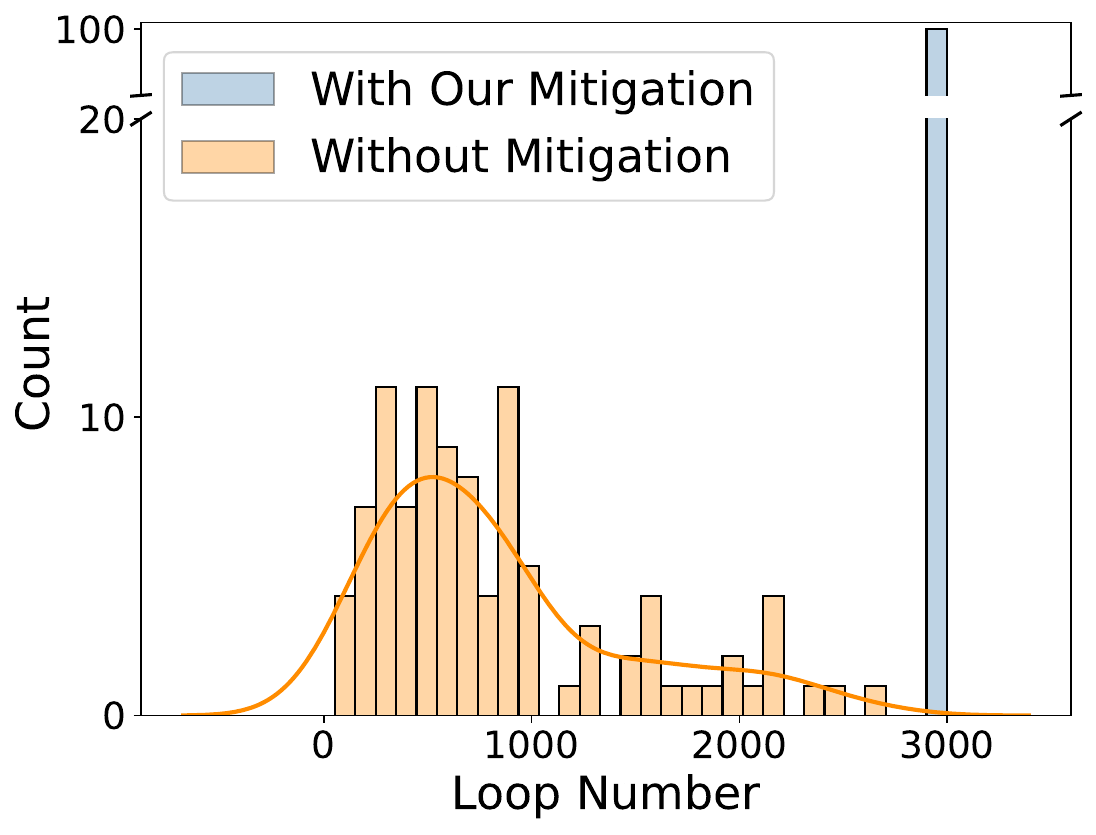}\vspace{-0.1cm}
    \caption{Distribution of loop numbers for attacks with and without our mitigation. Incorporating our mitigation increases attack costs, such that the attack is unsuccessful for all images after $3000$ rounds.}
    \label{loopnumber_with_wo_VAE}
\end{minipage}
\hfill
\begin{minipage}{0.32\textwidth}
    \centering
    \includegraphics[width=\linewidth]{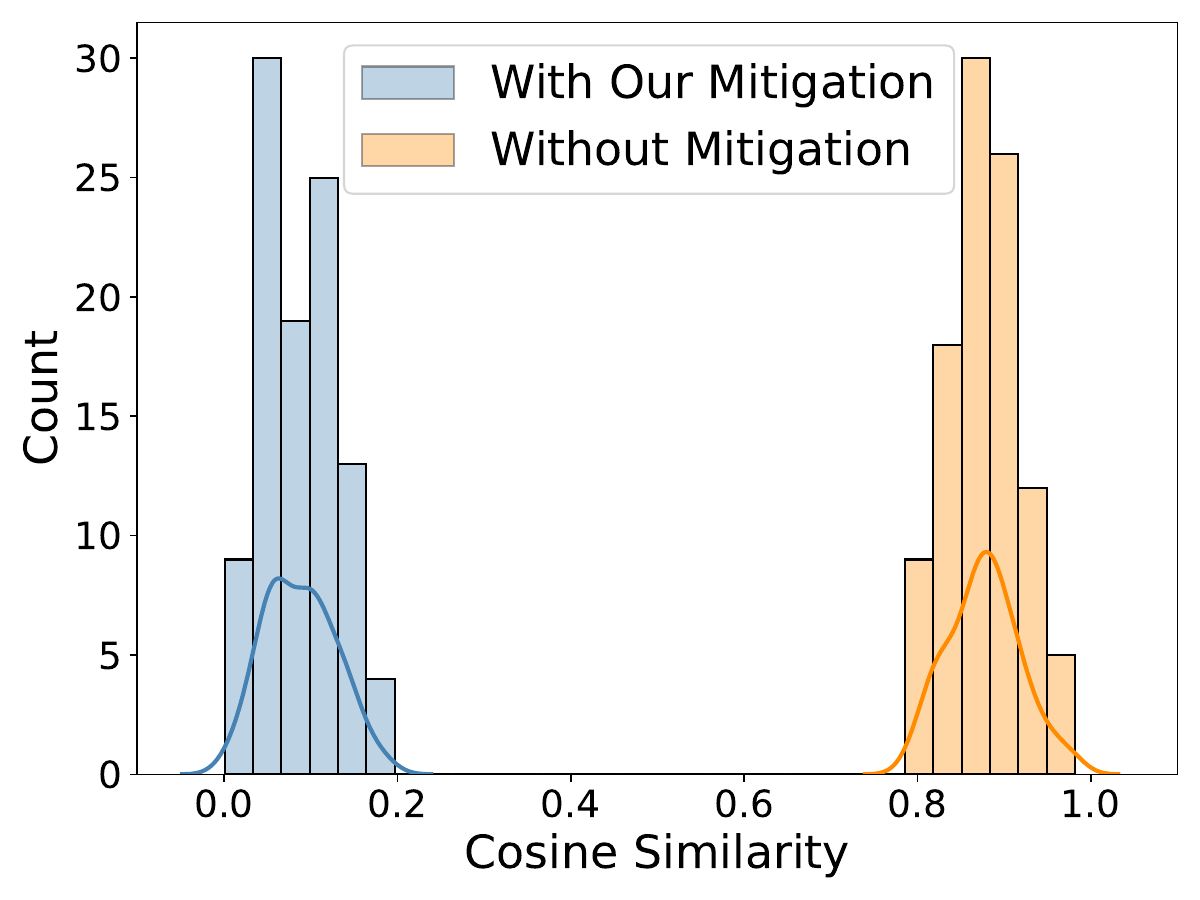}
    \caption{Distribution of cosine similarities between perturbed embeddings and target labels. Attacks with our mitigation yield low cosine values, whereas attacks without it reach the maximum similarity threshold.}
    \label{cosine_with_wo_VAE}
\end{minipage}

\end{figure*}

\subsection{Mitigation Computational Overheads}
Let us now discuss the trade-off between defense success rate and the computational overheads of our proposed defense mechanism.  Figure~\ref{sampling_effect} analyzes the effect of varying the number of generated samples used during the generative sampling stage for both \gls{VAE} and \gls{DM} models. As the number of generated samples increases, both Top-1 and Top-5 accuracies improve consistently for original and perturbed images. This indicates that aggregating over more samples provides a smoother and more reliable purification of the input manifold, enhancing robustness against adversarial illusions. However, the trend stabilizes after approximately $10$ samples, implying that beyond this point, additional sampling offers limited benefit relative to computational cost.
The computational overheads of the VAE and DM are $0.3\pm 0.02$ and $0.11\pm 0.18$ seconds, respectively, which are both considered negligible compared to the processing pipeline in multi-modal models (multi-modal encoder, downstream task, and classifier) with $18.28\pm 0.79$ seconds. This demonstrates that the overheads introduced by the VAE and DM are less than $2\%$ of the entire processing pipeline we consider here. The number of parameters also follow a similar pattern. In particular, the VAE has $83.61$ million parameters, which is marginal compared to the processing pipeline in multi-modal models with $2157.52$ million parameters.

\subsection{Attack Cost}

In this section, we examine whether an attacker who has access to the generative model used in our mitigation can incorporate it into the attack generation loop to achieve a successful attack, and we evaluate the associated attack cost.
The \textit{attack cost} quantifies the difficulty of generating an adversarial illusion, measured by the number of optimization steps or model queries required for a successful attack. A higher attack cost indicates that the adversary must spend more computational or query resources to align the perturbed embedding with the target label, implying stronger robustness of the model or defense mechanism. 

To measure attack cost, we quantify the number of optimization steps required to achieve a successful attack. First, we establish a criterion for determining when an attack should be considered successful. To do so, we conduct an experiment using the normal pipeline without mitigation and run the attack optimization loop until the reconstructed image’s Top-1 prediction from the classifier matches the target class. At that point, we record the \gls{cs} between the perturbed embedding and the target label.

The distribution of cosine similarities is shown in Fig.~\ref{cosine_exp}. This analysis allows us to identify a meaningful cutoff for when an attack can be considered successful, ensuring that comparisons of attack cost reflect genuine robustness rather than arbitrary stopping rules. The cosine similarities exhibit a mean of $0.52 \pm 0.12$. Based on this distribution, we adopt a conservative threshold of $0.8$, which covers approximately $98\%$ of successful attacks. We also consider a maximum loop budget of $3000$ iterations, as attacks that reach the threshold of $0.8$ consistently achieve top-1 accuracy for the target label. 
If an attack does not reach the \gls{cs} threshold within this loop count budget we record it as failed and assign the maximum loop count when reporting attack cost and plotting loop number histograms.

Using these thresholds, we evaluate attack cost under two settings in which attacks optimize through the \gls{VAE} or operate directly in pixel space to compare the robustness gained from introducing the generative model. Incorporating the \gls{VAE} into the attack loop increases the attack cost significantly. When the adversary optimizes through the \gls{VAE}, perturbations must pass through the generative bottleneck before being evaluated by the encoder, which disrupts the gradient flow and constrains updates to the image manifold. Consequently, the attack converges slowly or fails entirely, leading to a high attack cost and low \gls{cs} between the perturbed image and target label, e.g., most attacks reach the maximum of $3000$ iterations with \gls{cs} below $0.19$, as shown in Figures~\ref{loopnumber_with_wo_VAE} and~\ref{cosine_with_wo_VAE}. 
When the attack is applied directly in pixel space without the \gls{VAE}, gradients propagate unimpeded, resulting in faster convergence, low attack cost, and high \gls{cs} with the target label, generally converging in fewer than $1000$ iterations with \gls{cs} above $0.8$, which represents the maximum threshold. In our experiments, the model that includes the \gls{VAE} consistently reaches the maximum loop budget without achieving the \gls{cs} threshold, confirming that the generative constraint substantially limits the attack’s effectiveness.
Figure~\ref{cosine_with_wo_VAE} shows that when we continue the attack optimization loops for 3000 iterations, the \gls{cs} between the perturbed image and the target label diverges sharply between the two settings. Without mitigation, the similarity approaches the target, whereas with our mitigation, it remains far from it. This demonstrates that even when the attacker includes the VAE in the optimization loop, the attack fails to align the perturbed embedding with the target, indicating that our mitigation effectively prevents successful adversarial illusions.

\subsection{Extension to Text Downstream Task}
\begin{figure}[t]
\centering
\includegraphics[width=0.85\columnwidth]{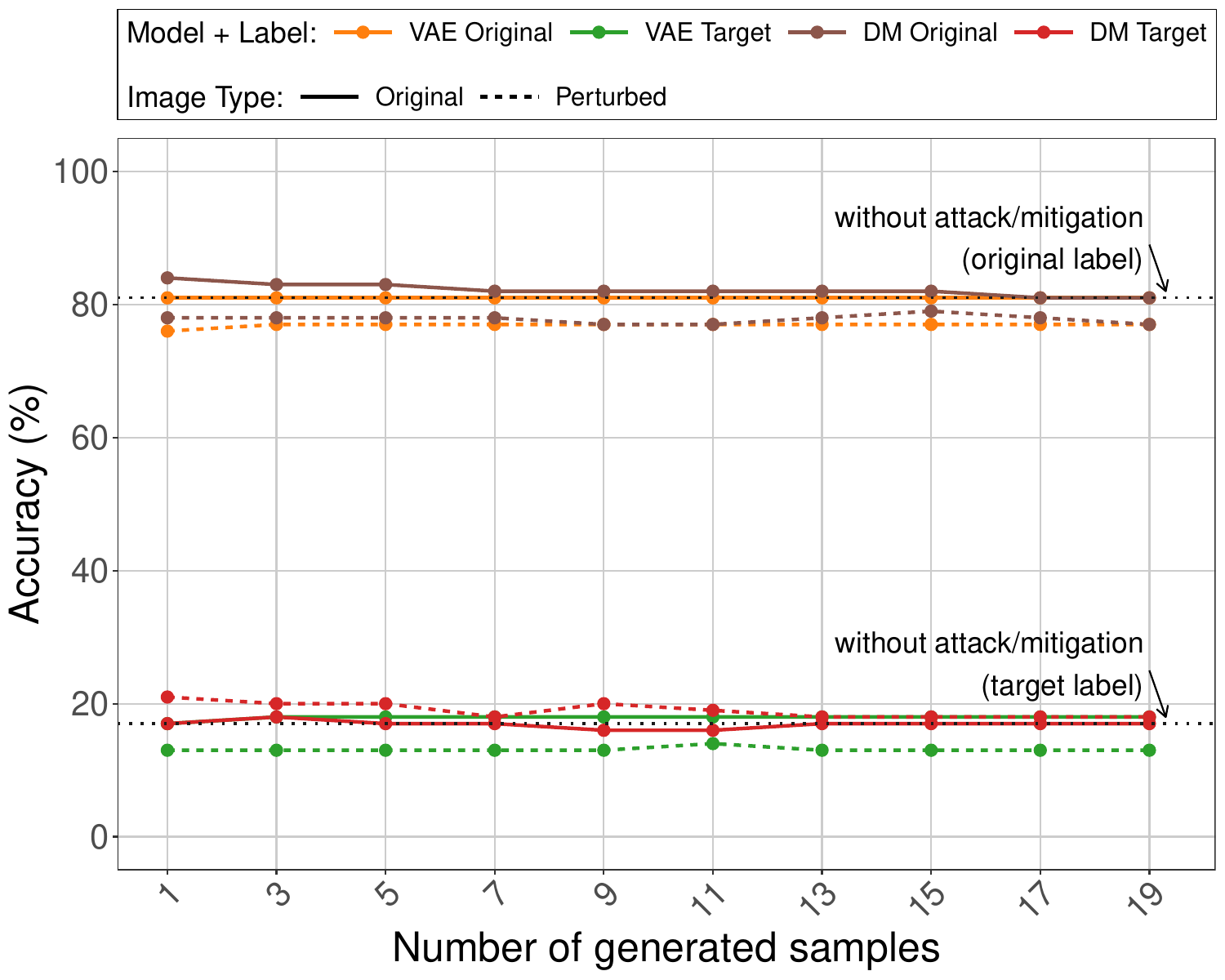}
\caption{Effect of sampling size on \emph{text generation downstream task} accuracy for \gls{VAE} and \gls{DM} mitigation methods. Dotted lines show baseline without attack/mitigation for the original and target labels. Accuracy for both original and target labels is reported across varying number of generated samples. Increasing the number of generated samples has a minor impact on text alignment accuracy.}
\label{text_generation}
\end{figure}

Figure~\ref{text_generation} presents the results of extending the adversarial illusion analysis to a text-generation downstream task. For each input image, either original or perturbed, we generate a single textual description. We then compute the similarity between the generated text embedding and both the original and target label embeddings using the \texttt{all-mpnet-base-v2} model from the Sentence-Transformers library~\cite{reimers2019sentence}, based on MPNet~\cite{song2020mpnet}.

The figure shows that across different numbers of generated samples, both VAE and DM maintain high similarity with the original label for clean images (solid lines), while adversarial perturbations (dashed lines) reduce this alignment slightly but consistently. Meanwhile, the similarity with the target label (green and red lines) remains low, indicating that the generative defenses largely prevent semantic transfer to the attacker-specified target. \gls{DM}-based purification (brown and red) achieves slightly higher stability than the VAE across all sampling sizes, demonstrating more consistent text alignment under adversarial conditions. Varying the number of generated samples shows no significant improvement, suggesting that a single generated sample per image is sufficient to capture the dominant semantic alignment for text-level evaluation.

\section{Related Work}

Multi-modal models integrate information from multiple modalities, such as images and text, to learn unified representations for cross-modal tasks~\cite{radford2021learning,yuan2021multimodal,srivastava2012multimodal}.
Unlike traditional adversarial perturbations that shift decision boundaries, adversarial illusions exploit cross-modal embedding misalignment, causing unrelated modalities to appear aligned in joint representation spaces~\cite{bagdasaryan2024adversarial}. For instance,~\citet{bagdasaryan2024adversarial} demonstrated that imperceptibly perturbed images could align with semantically unrelated text in models such as ImageBind and AudioCLIP, misleading both image and text generation pipelines without knowledge of the specific downstream tasks. Recent embedding-space and diffusion-guided attacks further highlight this vulnerability in vision-language models~\cite{shayegani2023jailbreak,kang2025trap}. Because adversarial illusions manipulate cross-modal alignment rather than decision boundaries, unimodal defenses that preserve label accuracy may still fail to maintain alignment across modalities, motivating generative approaches to restore shared embeddings. Thus, maintaining robust cross-modal alignment has emerged as a distinct challenge beyond unimodal adversarial robustness~\cite{zhou2023advclip,xia2025adversarial,Dou2024Adversarial}.

Beyond attacks that directly manipulate image–text embeddings, recent work has shown that adversarial images can reliably steer large multi-modal systems. ~\citet{wu2024adversarial} found that a single optimized trigger image can mislead web agents by forcing captioners or CLIP encoders to output attacker-chosen descriptions, which then propagate through downstream decision pipelines. A complementary line of work constructs universal adversarial images that jailbreak multi-modal LLMs across many prompts by optimizing through the vision–language stack~\cite{rahmatullaev2025universal}. These attacks assume gradient access to at least part of the multi-modal pipeline and target agent or language model behavior directly. Our setting differs because we evaluate purification-based defenses under a black-box setting, where the attacker cannot optimize through or modify the generative purifier.

Early heuristic defenses such as feature squeezing and JPEG-based transformations~\cite{xu2017feature,guo2017countering,das2018shield} inspired the use of autoencoders as learned purifiers. 
Convolutional \glspl{AE} have been widely studied as defenses against adversarial perturbations, aiming to remove input-space perturbations before classification~\cite{ashraf2024auto}. Early frameworks such as MagNet attempted to detect and purify adversarial samples but remained vulnerable to adaptive white-box attacks~\cite{meng2017magnet}. 
Denoising \glspl{AE} introduced robust latent representations~\cite{vincent2008extracting} and later outperformed traditional filtering for tasks like facial denoising~\cite{darici2023comparative}. Although effective for image restoration, these approaches have not been systematically evaluated under adversarial illusions, where the goal is preserving cross-modal alignment rather than unimodal fidelity.

\Glspl{VAE} extend \glspl{AE} with probabilistic latent spaces, enabling structured representations that can regularize noisy inputs~\cite{kingma2013auto}. The Adversarial Symmetric \gls{VAE} introduced adversarial training to improve purification and robustness~\cite{pu2017adversarial}, while more recent gated recurrent \glspl{VAE} have been used in industrial control systems for generating and defending against adversarial samples~\cite{xu2025gvae}. Defensive variants such as AdMVAE~\cite{yin2022defending} and Def-VAE~\cite{lu2023defvae} further demonstrate that probabilistic latent models can suppress adversarial signals in unimodal domains. However, the potential of these models for cross-modal alignment recovery, which is a key requirement in adversarial illusion scenarios, remains largely unexplored.

Generative \glspl{DM} have recently emerged as powerful tools for defending against adversarial attacks, particularly in multi-modal settings. By iteratively denoising inputs in a learned latent space, \glspl{DM} can restore both perceptual quality and semantic alignment~\cite{xia2025adversarial}. Frameworks such as DiffDf~\cite{diffdf2025}, ADBM~\cite{li2024adbm}, gradient-free purification~\cite{dai2025gradient}, and DiffCAP~\cite{fu2025diffcap} illustrate the growing use of diffusion-based purification for robustness. Compared to \gls{AE}- or \gls{VAE}-based methods, diffusion approaches provide greater flexibility and alignment recovery but incur higher computational cost. These advances suggest that generative purification may offer a promising path toward defending cross-modal representations from adversarial illusions.

Our consensus-based generative mitigation in this paper, by integrating generative models, stochastic sampling, and majority aggregation, reduces the adversarial illusions success rate to near-zero, while improving cross-modal alignment in unperturbed input settings without any attack/mitigation as well as in perturbed input settings compared to the existing solutions.

\section{Conclusions}
We provide a consensus-based generative mitigation against adversarial illusions in multi-modal models and demonstrate their efficacy on image and text downstream tasks. Since our defense requires no additional training, it can be applied readily in the state-of-the-art multi-modal models. Our experiments on the state-of-the-art multi-modal encoders show that our approach substantially reduces the attack success rates and improves cross-modal alignment, providing an effective and task-agnostic defense against adversarial illusions.

\section*{Acknowledgment}
This work is partially supported by the Wallenberg AI, Autonomous Systems and Software Program (WASP) funded by the Knut and Alice Wallenberg Foundation, the Swedish Research Council, and by an unrestricted gift from Google. The computations were enabled by resources provided by the National Academic Infrastructure for Supercomputing in Sweden (NAISS) partially funded by the Swedish Research Council (VR) through grant agreement no. 2022-06725.

\balance

{
    \small
    \bibliographystyle{ieeenat_fullname}
    \bibliography{main}

@inproceedings{zhou2023advclip,
  title={Advclip: Downstream-agnostic adversarial examples in multimodal contrastive learning},
  author={Zhou, Ziqi and Hu, Shengshan and Li, Minghui and Zhang, Hangtao and Zhang, Yechao and Jin, Hai},
  booktitle={Proceedings of the 31st ACM International Conference on Multimedia},
  pages={6311--6320},
  year={2023}
}

@article{xia2025adversarial,
  title={Adversarial-Guided Diffusion for Multimodal LLM Attacks},
  author={Xia, Chengwei and Ma, Fan and Quan, Ruijie and Zhan, Kun and Yang, Yi},
  journal={arXiv preprint arXiv:2507.23202},
  year={2025}
}

@inproceedings{bagdasaryan2024adversarial,
  title={Adversarial illusions in Multi-Modal embeddings},
  author={Zhang, Tingwei and Jha, Rishi and Bagdasaryan, Eugene and Shmatikov, Vitaly},
  booktitle={33rd USENIX Security Symposium (USENIX Security 24)},
  pages={3009--3025},
  year={2024}
}

@inproceedings{Dou2024Adversarial,
author = {Dou, Zhihao and Hu, Xin and Yang, Haibo and Liu, Zhuqing and Fang, Minghong},
title = {Adversarial Attacks to Multi-Modal Models},
year = {2024},
isbn = {9798400712098},
publisher = {Association for Computing Machinery},
address = {New York, NY, USA},
url = {https://doi.org/10.1145/3689217.3690619},
doi = {10.1145/3689217.3690619},
booktitle = {Proceedings of the 1st ACM Workshop on Large AI Systems and Models with Privacy and Safety Analysis},
pages = {35–46},
numpages = {12},
location = {Salt Lake City, UT, USA},
series = {LAMPS '24}
}

@article{ashraf2024auto,
  title={Auto encoder-based defense mechanism against popular adversarial attacks in deep learning},
  author={Ashraf, Syeda Nazia and Siddiqi, Raheel and Farooq, Humera},
  journal={PloS one},
  volume={19},
  number={10},
  pages={e0307363},
  year={2024},
  publisher={Public Library of Science San Francisco, CA USA}
}

@inproceedings{meng2017magnet,
  title={Magnet: a two-pronged defense against adversarial examples},
  author={Meng, Dongyu and Chen, Hao},
  booktitle={Proceedings of the 2017 ACM SIGSAC conference on computer and communications security},
  pages={135--147},
  year={2017}
}

@inproceedings{vincent2008extracting,
  title={Extracting and composing robust features with denoising autoencoders},
  author={Vincent, Pascal and Larochelle, Hugo and Bengio, Yoshua and Manzagol, Pierre-Antoine},
  booktitle={Proceedings of the 25th international conference on Machine learning},
  pages={1096--1103},
  year={2008}
}

@article{darici2023comparative,
  title={A comparative study on denoising from facial images using convolutional autoencoder},
  author={Dar{\i}c{\i}, Muazzez Buket and Erdem, Zeki},
  journal={Gazi University Journal of Science},
  volume={36},
  number={3},
  pages={1122--1138},
  year={2023},
  publisher={Gazi University}
}

@article{diffdf2025,
   title={Towards effective and efficient adversarial defense with diffusion models for robust visual tracking},
   volume={124},
   ISSN={1566-2535},
   url={http://dx.doi.org/10.1016/j.inffus.2025.103384},
   DOI={10.1016/j.inffus.2025.103384},
   journal={Information Fusion},
   publisher={Elsevier BV},
   author={Xu, Long and Gao, Peng and Tang, Wen-Jia and Wang, Fei and Yuan, Ru-Yue},
   year={2025},
   month=dec, pages={103384} }

@inproceedings{radford2021learning,
  title={Learning transferable visual models from natural language supervision},
  author={Radford, Alec and Kim, Jong Wook and Hallacy, Chris and Ramesh, Aditya and Goh, Gabriel and Agarwal, Sandhini and Sastry, Girish and Askell, Amanda and Mishkin, Pamela and Clark, Jack and others},
  booktitle={International conference on machine learning},
  pages={8748--8763},
  year={2021},
  organization={PmLR}
}

@inproceedings{yuan2021multimodal,
  title={Multimodal contrastive training for visual representation learning},
  author={Yuan, Xin and Lin, Zhe and Kuen, Jason and Zhang, Jianming and Wang, Yilin and Maire, Michael and Kale, Ajinkya and Faieta, Baldo},
  booktitle={Proceedings of the IEEE/CVF conference on computer vision and pattern recognition},
  pages={6995--7004},
  year={2021}
}

@article{srivastava2012multimodal,
  title={Multimodal learning with deep boltzmann machines},
  author={Srivastava, Nitish and Salakhutdinov, Russ R},
  journal={Advances in neural information processing systems},
  volume={25},
  year={2012}
}

@article{kingma2013auto,
  title={Auto-encoding variational bayes},
  author={Kingma, Diederik P and Welling, Max},
  journal={arXiv preprint arXiv:1312.6114},
  year={2013}
}

@article{xu2025gvae,
  title={G-VAE: Variational autoencoder-based adversarial attacks and defenses in industrial control systems},
  author={Xu, Lijuan and Yang, Zhi and Zhao, Dawei and Yu, Fuqiang and Zhou, Yang and Zhang, Hu},
  journal={Computers and Electrical Engineering},
  volume={124},
  pages={110290},
  year={2025},
  publisher={Elsevier}
}

@article{pu2017adversarial,
  title={Adversarial symmetric variational autoencoder},
  author={Pu, Yuchen and Wang, Weiyao and Henao, Ricardo and Chen, Liqun and Gan, Zhe and Li, Chunyuan and Carin, Lawrence},
  journal={Advances in neural information processing systems},
  volume={30},
  year={2017}
}

@article{shayegani2023jailbreak,
  title={Jailbreak in pieces: Compositional adversarial attacks on multi-modal language models},
  author={Shayegani, Erfan and Dong, Yue and Abu-Ghazaleh, Nael},
  journal={arXiv preprint arXiv:2307.14539},
  year={2023}
}

@article{kang2025trap,
  title={TRAP: Targeted Redirecting of Agentic Preferences},
  author={Kang, Hangoo and Yeon, Jehyeok and Singh, Gagandeep},
  journal={arXiv preprint arXiv:2505.23518},
  year={2025}
}

@article{li2024adbm,
  title={Adbm: Adversarial diffusion bridge model for reliable adversarial purification},
  author={Li, Xiao and Sun, Wenxuan and Chen, Huanran and Li, Qiongxiu and Liu, Yining and He, Yingzhe and Shi, Jie and Hu, Xiaolin},
  journal={arXiv preprint arXiv:2408.00315},
  year={2024}
}

@article{dai2025gradient,
  title={Gradient-free adversarial purification with diffusion models},
  author={Dai, Xuelong and Wang, Dong and Cheng, Xiuzhen and Xiao, Bin},
  journal={arXiv preprint arXiv:2501.13336},
  year={2025}
}

@article{fu2025diffcap,
  title={DiffCAP: Diffusion-based Cumulative Adversarial Purification for Vision Language Models},
  author={Fu, Jia and Wu, Yongtao and Chen, Yihang and Peng, Kunyu and Zhang, Xiao and Cevher, Volkan and Pashami, Sepideh and Holst, Anders},
  journal={arXiv preprint arXiv:2506.03933},
  year={2025}
}

@article{lu2023defvae,
  title={Def{VAE}: A defect detection method for catenary devices based on variational autoencoder},
  author={Lu, Tengfei and Wang, Zhongli and Shen, Yan and Shao, Xiaotao and Tang, Yonglin},
  journal={IEEE Transactions on Instrumentation and Measurement},
  volume={72},
  pages={1--12},
  year={2023},
  publisher={IEEE}
}

@article{yin2022defending,
  title={Defending against adversarial attacks using spherical sampling-based variational auto-encoder},
  author={Yin, Sheng-lin and Zhang, Xing-lan and Zuo, Li-yu},
  journal={Neurocomputing},
  volume={478},
  pages={1--10},
  year={2022},
  publisher={Elsevier}
}

@inproceedings{jia2021scaling,
  title={Scaling up visual and vision-language representation learning with noisy text supervision},
  author={Jia, Chao and Yang, Yinfei and Xia, Ye and Chen, Yi-Ting and Parekh, Zarana and Pham, Hieu and Le, Quoc and Sung, Yun-Hsuan and Li, Zhen and Duerig, Tom},
  booktitle={International conference on machine learning},
  pages={4904--4916},
  year={2021},
  organization={PMLR}
}

@inproceedings{girdhar2023imagebind,
  title={Imagebind: One embedding space to bind them all},
  author={Girdhar, Rohit and El-Nouby, Alaaeldin and Liu, Zhuang and Singh, Mannat and Alwala, Kalyan Vasudev and Joulin, Armand and Misra, Ishan},
  booktitle={Proceedings of the IEEE/CVF conference on computer vision and pattern recognition},
  pages={15180--15190},
  year={2023}
}

@article{goodfellow2014explaining,
  title={Explaining and harnessing adversarial examples},
  author={Goodfellow, Ian J and Shlens, Jonathon and Szegedy, Christian},
  journal={arXiv preprint arXiv:1412.6572},
  year={2014}
}

@article{madry2017towards,
  title={Towards deep learning models resistant to adversarial attacks},
  author={Madry, Aleksander and Makelov, Aleksandar and Schmidt, Ludwig and Tsipras, Dimitris and Vladu, Adrian},
  journal={arXiv preprint arXiv:1706.06083},
  year={2017}
}

@article{tramer2017ensemble,
  title={Ensemble adversarial training: Attacks and defenses},
  author={Tram{\`e}r, Florian and Kurakin, Alexey and Papernot, Nicolas and Goodfellow, Ian and Boneh, Dan and McDaniel, Patrick},
  journal={arXiv preprint arXiv:1705.07204},
  year={2017}
}

@inproceedings{athalye2018obfuscated,
  title={Obfuscated gradients give a false sense of security: Circumventing defenses to adversarial examples},
  author={Athalye, Anish and Carlini, Nicholas and Wagner, David},
  booktitle={International conference on machine learning},
  pages={274--283},
  year={2018},
  organization={PMLR}
}

@article{xu2017feature,
  title={Feature squeezing: Detecting adversarial examples in deep neural networks},
  author={Xu, Weilin and Evans, David and Qi, Yanjun},
  journal={arXiv preprint arXiv:1704.01155},
  year={2017}
}

@article{guo2017countering,
  title={Countering adversarial images using input transformations},
  author={Guo, Chuan and Rana, Mayank and Cisse, Moustapha and Van Der Maaten, Laurens},
  journal={arXiv preprint arXiv:1711.00117},
  year={2017}
}

@article{schlarmann2024robust,
  title={Robust clip: Unsupervised adversarial fine-tuning of vision embeddings for robust large vision-language models},
  author={Schlarmann, Christian and Singh, Naman Deep and Croce, Francesco and Hein, Matthias},
  journal={arXiv preprint arXiv:2402.12336},
  year={2024}
}

@inproceedings{das2018shield,
  title={Shield: Fast, practical defense and vaccination for deep learning using jpeg compression},
  author={Das, Nilaksh and Shanbhogue, Madhuri and Chen, Shang-Tse and Hohman, Fred and Li, Siwei and Chen, Li and Kounavis, Michael E and Chau, Duen Horng},
  booktitle={Proceedings of the 24th ACM SIGKDD International Conference on Knowledge Discovery \& Data Mining},
  pages={196--204},
  year={2018}
}

@article{reimers2019sentence,
  title={Sentence-bert: Sentence embeddings using siamese bert-networks},
  author={Reimers, Nils and Gurevych, Iryna},
  journal={arXiv preprint arXiv:1908.10084},
  year={2019}
}

@article{song2020mpnet,
  title={Mpnet: Masked and permuted pre-training for language understanding},
  author={Song, Kaitao and Tan, Xu and Qin, Tao and Lu, Jianfeng and Liu, Tie-Yan},
  journal={Advances in neural information processing systems},
  volume={33},
  pages={16857--16867},
  year={2020}
}

@article{wu2024adversarial,
  title={Dissecting adversarial robustness of multimodal lm agents},
  author={Wu, Chen Henry and Shah, Rishi and Koh, Jing Yu and Salakhutdinov, Ruslan and Fried, Daniel and Raghunathan, Aditi},
  journal={arXiv preprint arXiv:2406.12814},
  year={2024}
}

@article{rahmatullaev2025universal,
  title={Universal Adversarial Attack on Aligned Multimodal LLMs},
  author={Rahmatullaev, Temurbek and Druzhinina, Polina and Kurdiukov, Nikita and Mikhalchuk, Matvey and Kuznetsov, Andrey and Razzhigaev, Anton},
  journal={arXiv preprint arXiv:2502.07987},
  year={2025}
}

}

\clearpage
\setcounter{page}{1}
\maketitlesupplementary
\setcounter{section}{0}  
\renewcommand{\thesection}{A.\arabic{section}}

\nobalance
\section{Cosine Similarity Analysis of Mitigation Methods}
Table \ref{tab:table1_appendix} reports the cosine similarity between the embedding vectors of original/perturbed images and the corresponding outputs produced by each mitigation method. A high cosine similarity for the original input indicates low distortion or side effects, whereas a low cosine similarity for the perturbed input suggests that the mitigation method effectively disrupts the illusion attack by removing or weakening the adversarial signal.

\begin{table}[b]
\centering
\caption{Cosine similarity between original/perturbed images and their augmented versions. Here, $x$ denotes the original image and $\tilde{x}$ its perturbed counterpart. \text{Method}($\cdot$) represents the transformation produced by a given mitigation technique. \text{CS}($\cdot,\cdot$) computes the cosine similarity between the embedding vectors of two inputs.}
\resizebox{\columnwidth}{!}{
\begin{tabular}{lcc}
\hline
\textbf{Method} &
\textbf{CS($x$, Method($x$))} &
\textbf{CS($\tilde{\vx}$, Method($\tilde{\vx}$))} \\
\hline
VAE + Sampling (Ours) & $0.93 \pm 0.02$ & $0.15 \pm 0.05$ \\
DM + Sampling (Ours) & $0.92 \pm 0.04$ & $0.15 \pm 0.05$ \\
JPEG & $0.85 \pm 0.04$ & $0.15 \pm 0.05$ \\
Gaussian Blur & $0.77 \pm 0.05$ & $0.13 \pm 0.04$ \\
Random Affine & $0.84 \pm 0.05$ & $0.15 \pm 0.04$ \\
Color Jitter & $0.88 \pm 0.04$ & $0.28 \pm 0.07$ \\
Horizontal Flip & $0.90 \pm 0.03$ & $0.26 \pm 0.10$ \\
Random Perspective & $0.80 \pm 0.06$ & $0.16 \pm 0.04$ \\
\hline
\end{tabular}
}
\label{tab:table1_appendix}
\end{table}

Among all evaluated approaches, VAE + Sampling exhibits the most favorable trade-off. It achieves the highest cosine similarity on original images (0.93), indicating that it preserves the clean content more faithfully than the other methods. At the same time, its similarity on perturbed images drops to 0.15, meaning that the filtered output is largely unaligned with the adversarially manipulated representation and thus effectively disrupts the attacker’s intended semantics.

Other mitigation methods achieve comparable cosine similarity values on the perturbed inputs, indicating that they can also diminish the adversarial manipulation to some extent. However, these methods generally introduce stronger distortions to clean images. For example, Gaussian Blur yields the lowest cosine similarity on perturbed images (0.13), suggesting effective removal of the illusion signal, but it also reduces the similarity for clean images to 0.77—substantially lower than the preservation achieved by VAE + Sampling. This highlights that while several methods can weaken the illusion attack, VAE + Sampling offers the best balance between attack mitigation and fidelity to the original image.

\begin{table}[t]
\centering
\caption{Computational cost of VAE, Diffusion and Original Models in terms of parameters and FLOPs.}
\begin{tabular}{lcc}
\hline
\textbf{Method} & \textbf{Parameters (M)} & \textbf{TFLOPs} \\
\hline
VAE & $83.65$ & $3.57$ \\
DM & $859.52$ & $33.9$ \\
Original & $2157.52$ & $155.6$ \\
\hline
\end{tabular}
\label{tab:table2_comparison}
\end{table}

\section{Computational Cost Analysis of Mitigation Methods}
Table~\ref{tab:table2_comparison} presents the parameters and FLOPs of the original adversarial illusion model and the VAE and diffusion mitigation modules. The originsl model contains 2,157.52M parameters and requires 155.6 TFLOPs for one image. The VAE mitigation is highly lightweight, adding only 83.65M parameters and 3.57 TFLOPs, which corresponds to merely a 3.9\% increase in parameters and a 2.3\% increase in computation over the original model. The diffusion model is heavier with a 39.8\% parameter and 21.8\% FLOP overhead compared to original model. Both mitigation modules introduce only a relatively small extra computational cost.

\end{document}